# Domain Filtering Consistencies

**Romuald Debruyne**                                    Romuald.Debruyne@emn.fr
*Member of the* Coconut *group*
*Ecole des Mines de Nantes,*
*La Chantrerie, 4, Rue Alfred Kastler, 44307 Nantes Cedex 3 - France*

**Christian Bessière**                                    bessiere@lirmm.fr
*Member of the* Coconut *group*
*LIRMM - CNRS UMR 5506, 161 rue Ada, 34392 Montpellier Cedex 5 - France*

## Abstract

Enforcing local consistencies is one of the main features of constraint reasoning. Which level of local consistency should be used when searching for solutions in a constraint network is a basic question. Arc consistency and partial forms of arc consistency have been widely studied, and have been known for sometime through the forward checking or the MAC search algorithms. Until recently, stronger forms of local consistency remained limited to those that change the structure of the constraint graph, and thus, could not be used in practice, especially on large networks. This paper focuses on the local consistencies that are stronger than arc consistency, without changing the structure of the network, i.e., only removing inconsistent values from the domains. In the last five years, several such local consistencies have been proposed by us or by others. We make an overview of all of them, and highlight some relations between them. We compare them both theoretically and experimentally, considering their pruning efficiency and the time required to enforce them.

## 1. Introduction

There are more and more applications in artificial intelligence that use constraint networks (CNs) to solve combinatorial problems, ranging from design to diagnosis, resource allocation to car sequencing, natural language understanding to machine vision. Finding a solution in a constraint network involves looking for a set of value assignments, one for each variable, so that all the constraints are simultaneously satisfied (Meseguer, 1989; Tsang, 1993). This task is NP-hard and many exponential time algorithms have been proposed to solve this problem. These algorithms, which make a systematic exploration of the search space, all have backtracking as a basis. As long as the unassigned variables have values consistent with the partial instantiation, they extend it by assigning values to variables. Otherwise, a dead-end is reached and some previous assignments have to be changed before going on with the partial instantiation extension. The explicit constraints of the network together induce some implicit constraints. Since basic search algorithms do not record these implicit constraints, they waste time by repeatedly detecting the local inconsistencies caused by them. Filtering techniques are essential to reduce the size of the search space and so to improve the efficiency of search algorithms. They can be used during a preprocessing step to remove once and for all some local inconsistencies that otherwise would have been repeatedly found during search (Dechter & Meiri, 1994). They can also be maintained during search.

Search algorithms differ in the kind of local consistency they achieve after each choice of a value for a variable. Most of them enforce partial arc consistency, going from forward checking (FC,Golomb & Baumert, 1965; Haralick & Elliott, 1980), which only removes the values directly arc inconsistent with the last assignment, to really full look-ahead (RFL, Gaschnig, 1974), which achieves full arc consistency. Arc consistency (AC) and partial forms of arc consistency are widely used for two reasons. First, they have low space and time complexities, while path consistency and higher levels of $k$-consistency, which were for a long time the only other options, are prohibitive and can change the structure of the network. Moreover, until 1995, more pruningful local consistencies seemed uninteresting since experimental evaluations of search algorithms showed that the limited local consistency used by forward checking was the best choice (Nadel, 1988; Kumar, 1992; Bacchus & van Run, 1995). This conclusion is not surprising since the comparisons were made on very small and easy constraint networks. On such problems, the additional cost of pruning more values could not be outweighed by the search savings.

However, the harder a constraint network is, the more useful filtering techniques are. More recent works (Bessière & Régin, 1996; Sabin & Freuder, 1994; Grant & Smith, 1996) testing search algorithms at the threshold (Cheeseman, Kanefsky, & Taylor, 1991), where most of the hard problems are expected to be, show that using more pruningful local consistencies can be worthwhile. Their conclusion is that maintaining arc consistency during search (MAC), namely achieving AC both after the choice of a value for a variable and after the refutation of such a choice, outperforms forward checking on hard problems. The good behavior of MAC is even more significant on large problems, especially when domains are large. It is conceivable that on very difficult instances, maintaining an even more pruningful local consistency may pay off. Obviously, such an algorithm would waste seconds on easy CNs but it would save many minutes (hours ?) on very hard problems, reducing the set of pathological CNs on which search is really prohibitive.

In this paper we study the local consistencies as preprocessing filtering techniques. This is a preliminary work before trying to determine which local consistency is the most advantageous to be maintained during search. In the last five years, many new local consistencies have been proposed. In the remaining of this paper, we focus our attention on those that leave unchanged the structure of the network since they are the only able to be used on large CNs. In addition to an overview of these local consistencies that only remove inconsistent values, we both compare, theoretically and experimentally, their pruning efficiency and the time needed to enforce them.

## 2. Definitions and Notations

A *network of binary constraints* $P = (\mathcal{X}, \mathcal{D}, \mathcal{C})$ is defined by a set $\mathcal{X} = \{i, j, \dots\}$ of $n$ *variables*, each taking value in its respective finite *domain* $D_i, D_j, \dots$ elements of $\mathcal{D}$, and a set $\mathcal{C}$ of $e$ binary constraints. $d$ is the size of the largest domain. A *binary constraint* $C_{ij}$ is a subset of the Cartesian product $D_i \times D_j$ that denotes the compatible pairs of values for $i$ and $j$. We denote $C_{ij}(a, b) = true$ to specify that $((i, a), (j, b)) \in C_{ij}$. We then say that $(j, b)$ is a *support* for $(i, a)$ on $C_{ij}$. Checking whether a pair of values is allowed by a constraint is called a *constraint check*. With each CN we associate a *constraint graph* in which nodes represent variables and arcs connect pairs of variables that are constrained explicitly. $c$ is the number of 3-cliques in the constraint graph and $g$ is the maximum degree of a node in the constraint graph. The *neighborhood* of $i$ is the set of variables adjacent to $i$ in the constraint graph. A domain $\mathcal{D}' = \{D_i', D_j', \ldots\}$ is a *sub-domain* of $\mathcal{D} = \{D_i, D_j, \ldots\}$ if $\forall i, D_i' \subseteq D_i$. An *instantiation* $I$ of a set of variables $S$ is a set of value assignments $\{I_j\}_{j \in S}$, one for each variable belonging to $S$, s.t. $\forall j \in S$, $I_j \in D_j$. An instantiation $I$ of $S$ satisfies a constraint $C_{ij}$ if $\{i, j\} \not\subseteq S$ or $C_{ij}(I_i, I_j)$ is true. An instantiation is *consistent* if it satisfies all the constraints. A pair of values $((i, a), (j, b))$ is *path consistent* if for all $k \in \mathcal{X}$ s.t. $j \neq k \neq i \neq j$, this pair of values can be extended to a consistent instantiation of $\{i, j, k\}$. $(j, b)$ is a *path consistent support* for $(i, a)$ if $(a, b) \in C_{ij}$ and $((i, a), (j, b))$ is path consistent. A *solution* of $P = (\mathcal{X}, \mathcal{D}, \mathcal{C})$ is a consistent instantiation of $\mathcal{X}$. A value $(i, a)$ is *consistent* if there is a solution $I$ such that $I_i = a$, and a CN is *consistent* if it has at least one solution. We denote by $P|_{D_i = \{a\}}$ the CN obtained by restricting $D_i$ to $\{a\}$ in $P$. If $LC$ is a local consistency, a CN $P$ is not *LC-consistent* iff $LC$ does not hold on $P$. A CN $P$ is *LC-inconsistent* iff we cannot obtain a *LC*-consistent constraint network by deletion of some local inconsistencies in $P$. If a local consistency $LC$ is used to detect the inconsistency of instantiations no longer than 1, we can say that a CN $P = (\mathcal{X}, \mathcal{D}, \mathcal{C})$ is *LC-inconsistent* iff there is no sub-domain $\mathcal{D}'$ of $\mathcal{D}$ such that $LC$ holds on $(\mathcal{X}, \mathcal{D}', \mathcal{C})$.

## 3. The Local Consistencies Studied

Filtering techniques can be used to detect the inconsistency of a CN, and under some assumptions, they can ensure a backtrack-free search (Freuder, 1982, 1985). However, their usual purpose is not to find a solution in a constraint network. They remove some local inconsistencies and so delete some regions of the search space that do not contain any solution. The resulting CN is equivalent to the initial one since the set of solutions is unchanged, but if substantial reductions are made the search becomes easier. In this section we review the basis of arc consistency, $k$-consistency, restricted path consistency, and inverse consistencies. Furthermore, we extend the idea of restricted path consistency to $k$-restricted path consistency and Max-restricted path consistency. We propose a new class of local consistencies called singleton consistencies. We also show a property of path inverse consistency that can be used to have an optimal worst case time complexity in a path inverse consistency algorithm (Debruyne, 2000).

**Arc consistency** The most widely used local consistency is arc consistency. It is based on the notion of support. A value is viable as long as it has at least one compatible value in the domain of each neighboring variable. An AC algorithm has to remove the values that have no support on a constraint. As in most of the filtering techniques, the value deletions have to be propagated through the network since they can lead to the arc inconsistency of some values that were previously viable.

**$k$-consistency** A consistent instantiation of length $k$-1 is $k$-consistent (i.e., $(k$-1, 1)-consistent in the formalism of Freuder, 1985) if it can be extended to any additional $k^{th}$ variable. The time and space complexities of enforcing $k$-consistency are polynomial with the exponent dependent on $k$. If $k \geq 3$, the constraints have to be represented in extension to store the $(k$-1)-tuples deleted, and the structure of the network can be changed. This leads to huge space requirements and subsequently important cpu time costs. In practice, only 2-consistency, which is arc consistency in binary CNs, can be used. Although path consistency (PC, namely 3-consistency) cannot be used on large CNs, our experiments will involve *strong* path consistency (namely enforcing both arc and path consistency) because PC has been widely studied.

**Restricted path consistency** The aim of Berlandier when he proposed restricted path consistency (RPC, Berlandier, 1995) was to remove more inconsistent values than AC while avoiding the drawbacks of path consistency. Even the most efficient PC algorithms have to try to extend all the pairs of values (even those between two variables that are not neighbors) to any third variable. The basis of RPC is to avoid most of this prohibitive work. RPC performs only the most pruningful path consistency checks, namely those able to directly delete a value. In addition to AC, an RPC algorithm checks the path consistency of the pairs of values $((i,a),(j,b))$ such that $(j,b)$ is the only support for $(i,a)$ in $D_j$. If such a pair is path inconsistent, its deletion would lead to the arc inconsistency of $(i,a)$. Thus $(i,a)$ can be removed. These few additional path consistency checks allow detecting more inconsistent values than AC without having to delete any pair of values, and so leaving unchanged the structure of the network.

**$k$-restricted path consistency** We can extend the idea of RPC to a more pruningful local consistency. If RPC holds, the values that have only one support on a constraint are such that this support is path consistent. Checking the path consistency of more supports can remove even more values without falling into the traps of PC. $k$-restricted path consistency ($k$-RPC, Debruyne & Bessière, 1997a) looks for a path consistent support on a constraint for the values that have at most $k$ supports on this constraint. RPC is 1-RPC and AC corresponds to 0-RPC. If $k$-RPC holds, to achieve $(k+1)$-RPC we only have to check the values that have exactly $(k+1)$ supports on a constraint and to propagate their possible deletion. So, it is possible to achieve AC, RPC, 2-RPC and so on, each time reusing previous filtering effort. This adaptive enforcement can be stopped as soon as each value has at least one path consistent support on each constraint, the constraint network being $d$-RPC where $d$ is the size of the largest domain. Indeed, if after achieving $k$-RPC all the values have at most $k$ supports on each constraint, $k'$-RPC holds for all $k' > k$.

**Max-restricted path consistency** A constraint network is Max-restricted path consistent (Max-RPC, Debruyne & Bessière, 1997a) if all the values have at least one path consistent support on each constraint, whatever is the number of supports. From the pruning efficiency point of view, Max-RPC is an upper bound for $k$-RPC. Achieving Max-RPC involves deleting all the $k$-restricted path inconsistent values for all $k$. However, achieving Max-RPC can be less expensive than enforcing a high level of $k$-RPC. As opposed to Max-RPC, to achieve $k$-RPC we have to look for the values that have at most $k$ supports on a constraint to know the values for which a path consistent support has to be found. This can be expensive if $k$ is great, the algorithm having to look for $k+1$ supports for each value on each constraint. Unconditionally looking for a path consistent support avoids this costly extra work.

**$k$ inverse consistency** The aim of Freuder and Elfe when they proposed inverse consistency (Freuder & Elfe, 1996) was to achieve high order local consistencies with a good space complexity. A $k$-consistency algorithm removes the instantiation of length $k-1$ that cannot be extended to any $k^{th}$ variable. It requires $O(n^{k-1}d^{k-1})$ space to keep track of the deleted instantiations. Space requirements are no longer a problem with $k$ inverse consistency $((1, k\text{-}1)$-consistency), which removes the values that cannot be extended to a consistent instantiation including any $k$-1 additional variables. Its linear space complexity would allow using it on large CNs. However, its worst case time complexity is polynomial with the exponent dependent on $k$, which restricts its use to small values of $k$.

**Path inverse consistency** The first level of $k$ inverse consistency removing more values than AC is path inverse consistency (PIC, $k = 3$). By definition, $(i, a)$ is path inverse consistent if it can be extended to all the 3-tuples of variables containing $i$. However, as said in (Freuder & Elfe, 1996), not all the 3-tuples need to be checked to enforce PIC. Only one of the tuples $(i, j, k)$ and $(i, k, j)$ has to be checked. Moreover, if $i$ is linked to neither $j$ nor $k$, $(i, a)$ can be deleted because of $(i, j, k)$ only if all the values of $j$ or $k$ are path inverse inconsistent. In such a case, checking $(i, j, k)$ is useless since PIC detects the inconsistency of the network when processing $j$ or $k$.

**Neighborhood inverse consistency** Since $k$ inverse consistency is polynomial with the exponent dependent on $k$, checking the $k$ inverse consistency of all the values is prohibitive if $k$ is great. However, if the variables are not uniformly constrained, it would be worthwhile to adapt the level of $k$ inverse consistency to the number of constraints involving them, focusing filtering effort on the most constrained variables (as it is done for adaptive consistency Dechter & Pearl, 1988). This is the basis of neighborhood inverse consistency (NIC, Freuder & Elfe, 1996), which removes the values that cannot be extended to a consistent instantiation including all the neighboring variables. We have to point out that the behavior of NIC is dependent on the structure of the constraint graph. If two variables $i$ and $j$ are not neighbors, we can add a universal constraint allowing all the pairs of values $(a, b) \in D_i \times D_j$ between $i$ and $j$. The resulting CN is equivalent to the initial one since it has the same set of solutions. However, as opposed to the other filterings studied in this paper, NIC is affected by this change since it can remove more values. Obviously, this process increases time complexity. On complete constraint networks, NIC tries to extend all the values to a whole solution, namely deleting all the globally inconsistent values (named variable completability in Freuder, 1991). This is a far more difficult task than looking for one solution. To be cost effective, NIC has to be used only on sparse CNs, where the degree of the variables is small.

**Singleton consistencies** If a value $(i, a)$ is consistent, the constraint network obtained by restricting $D_i$ to the singleton $\{a\}$ is consistent. Singleton consistencies are a class of filtering techniques based on this remark. To detect the inconsistency of a value $(i, a)$, a singleton consistency filtering technique checks whether a given local consistency can detect the possible inconsistency of $P|_{D_i=\{a\}}$. For example, singleton arc consistency (SAC, Debruyne & Bessière, 1997b) deletes the values $(i, a)$ such that $P|_{D_i=\{a\}}$ has no arc consistent sub-domain.[1] SAC has been inspired by the strong path consistency algorithm proposed by McGregor (McGregor, 1979). A SAC algorithm is obtained by omitting the deletions

---

1. Any AC algorithm can be used to know whether enforcing AC on $P|_{D_i=\{a\}}$ leads to a domain wipe out, but a lazy approach (such as LAC7 Schiex, Régin, Gaspin, & Verfaillie, 1996) is sufficient.

- A binary CN is $(i, j)$-*consistent* iff $\forall i \in \mathcal{X}$, $D_i \neq \emptyset$ and any consistent instantiation of $i$ variables can be extended to a consistent instantiation including any $j$ additional variables.

- A value $a \in D_i$ is arc consistent iff, $\forall j \in \mathcal{X}$ s.t. $C_{ij} \in \mathcal{C}$, there exists $b \in D_j$ s.t. $C_{ij}(a, b)$. A domain $D_i$ is arc consistent iff, $\forall a \in D_i$, $(i, a)$ is arc consistent. A CN is *arc consistent* ($(1, 1)$-consistent) iff $\forall D_i \in \mathcal{D}$, $D_i$ is a non empty arc consistent domain.

- A pair of values $((i, a), (j, b))$ is *path consistent* iff $\forall k \in \mathcal{X}$, there exists $c \in D_k$ s.t. $C_{ik}(a, c)$ and $C_{jk}(b, c)$, otherwise it is path inconsistent. A CN is *path consistent* ($(2, 1)$-consistent) iff no path inconsistent pair of values is allowed.

- A binary CN is *strongly path consistent* iff it is node consistent, arc consistent and path consistent.

- A binary CN is *path inverse consistent* iff it is $(1, 2)$-consistent i.e., $\forall (i, a) \in \mathcal{D}$  $\forall j, k \in \mathcal{X}$ s.t.  $j \neq i \neq k \neq j$, $\exists (j, b) \in \mathcal{D}$ and $\exists (k, c) \in \mathcal{D}$ s.t. $C_{ij}(a, b) \wedge C_{ik}(a, c) \wedge C_{jk}(b, c)$

- A binary CN is *neighborhood inverse consistent* iff $\forall (i, a) \in \mathcal{D}$, $(i, a)$ can be extended to a consistent instantiation including the neighborhood of $i$.

- A binary CN is *restricted path consistent* iff
  $\forall i \in \mathcal{X}$, $D_i$ is a non empty arc consistent domain and,
  $\forall (i, a) \in \mathcal{D}$, for all $j \in \mathcal{X}$ s.t. $(i, a)$ has an unique support $b$ in $D_j$,
  for all $k \in \mathcal{X}$ linked to both $i$ and $j$, $\exists c \in D_k$ s.t. $C_{ik}(a, c) \wedge C_{jk}(b, c)$.

- A binary CN is *k-restricted path consistent* iff
  $\forall i \in \mathcal{X}$, $D_i$ is a non empty arc consistent domain and,
  $\forall (i, a) \in \mathcal{D}$, for all $C_{ij} \in \mathcal{C}$ s.t. $(i, a)$ has at most $k$ supports in $D_j$,
  $\exists b \in D_j$ s.t. $C_{ij}(a, b)$ and
  $\forall k \in \mathcal{X}$ linked to both $i$ and $j$, $\exists c \in D_k$ s.t. $C_{ik}(a, c) \wedge C_{jk}(b, c)$.

- A binary CN is *max-restricted path consistent* iff
  $\forall i \in \mathcal{X}$, $D_i$ is a non empty arc consistent domain and,
  $\forall (i, a) \in \mathcal{D}$, for all $C_{ij} \in \mathcal{C}$,
  $\exists b \in D_j$ s.t. $C_{ij}(a, b)$ and
  $\forall k \in \mathcal{X}$ linked to both $i$ and $j$, $\exists c \in D_k$ s.t. $C_{ik}(a, c) \wedge C_{jk}(b, c)$.

- A binary CN $P$ is *singleton arc consistent* iff $\forall i \in \mathcal{X}$, $D_i \neq \emptyset$ and $\forall (i, a) \in \mathcal{D}$, $P|_{D_i = \{a\}}$ has an arc consistent sub-domain.

- A binary CN $P$ is *singleton restricted path consistent* iff $\forall i \in \mathcal{X}$, $D_i \neq \emptyset$ and $\forall (i, a) \in \mathcal{D}$, $P|_{D_i = \{a\}}$ has a restricted path consistent sub-domain.

Figure 1: The mentioned local consistencies.

| Name of<br>the algorithm | Worst case<br>time complexity | Worst case<br>space complexity |
|---|---|---|
| AC7 (Bessière, Freuder, & Régin, 1999) | $O(ed^2)^{(*)}$ | $O(ed)$ |
| RPC2 (Debruyne & Bessière, 1997a) | $O(en + ed^2 + cd^2)^{(*)}$ | $O(ed + cd)$ |
| Max RPC1 (Debruyne & Bessière, 1997a) | $O(en + ed^2 + cd^3)^{(*)}$ | $O(ed + cd)$ |
| PC5 (Singh, 1995)<br>PC8 (Chmeiss & Jégou, 1996) | $O(n^3d^3)^{(*)}$<br>$O(n^3d^4)$ | $O(n^3d^2)$<br>$O(n^2d)^2$ |
| PIC1 (Freuder & Elfe, 1996)<br>PIC2 (Debruyne, 2000) | $O(en^2d^4)$<br>$O(en + ed^2 + cd^3)^{(*)}$ | $O(n)$<br>$O(ed + cd)$ |
| NIC1 (Freuder & Elfe, 1996) | $O(g^2(n + ed)d^{g+1})$ | $O(n)$ |
| SAC1 (Debruyne & Bessière, 1997b) | $O(en^2d^4)$ | $O(ed)$ |
| SRPC1 (Debruyne & Bessière, 1997b) | $O(en + n^2(e + c)d^4)$ | $O(ed + cd)$ |

$^{(*)}$ optimal worst case time complexity.

Table 1: The most efficient algorithms achieving the mentioned local consistencies.[3]

of pairs of values in that algorithm. Many other singleton consistencies can be considered since any local consistency can be used to detect the possible inconsistency of the CNs $P|_{D_i = \{a\}}$ with $(i, a) \in \mathcal{D}$. If a local consistency can be enforced in a polynomial time, the corresponding singleton consistency also has a polynomial worst case time complexity.

The formal definitions of the local consistencies studied in this paper are presented in Figure 1. Table 1 recalls the time and space complexities of the most efficient algorithms enforcing them. The worst case time complexity of SAC1, SRPC1, and NIC1 have not been proved to be optimal.

## 4. Relations between PIC, RPC and Max-RPC

To highlight the relations between PIC, RPC and Max-RPC, let us show a property of path inverse consistency. As shown in (Debruyne, 2000), if we assume that the constraint network is arc consistent, enforcing PIC requires checking even less 3-tuples than those mentioned in (Freuder & Elfe, 1996). If $(i, a)$ is arc consistent, it can be extended to any 3-tuple $(i, j, k)$ such that there is no constraint between $j$ and $k$. Indeed, $(i, a)$ has a support $(j, b)$ on $C_{ij}$ and a support $(k, c)$ on $C_{ik}$, and since $j$ is not linked to $k$, $((i, a), (j, b), (k, c))$ is consistent. Furthermore, $(i, a)$ can be extended to $(i, j, k)$ if there is no constraint between $i$ and $k$ (resp. between $i$ and $j$). Indeed, $(i, a)$ has a support $b$ in $D_j$ (resp. $c$ in $D_k$) and this value being arc consistent too, it has a support $c$ in $D_k$ (resp. $b$ in $D_j$). So, $((i, a), (j, b), (k, c))$ is consistent. Consequently, if the constraint network is arc consistent, the only 3-tuples that have to be checked to achieve PIC correspond to the 3-cliques of the constraint graph.

---

2. However a $O(n^2d^2)$ data structure is required for the constraint representation.

3. See Section 2 for a definition of $n$, $d$, $e$, $c$, and $g$.

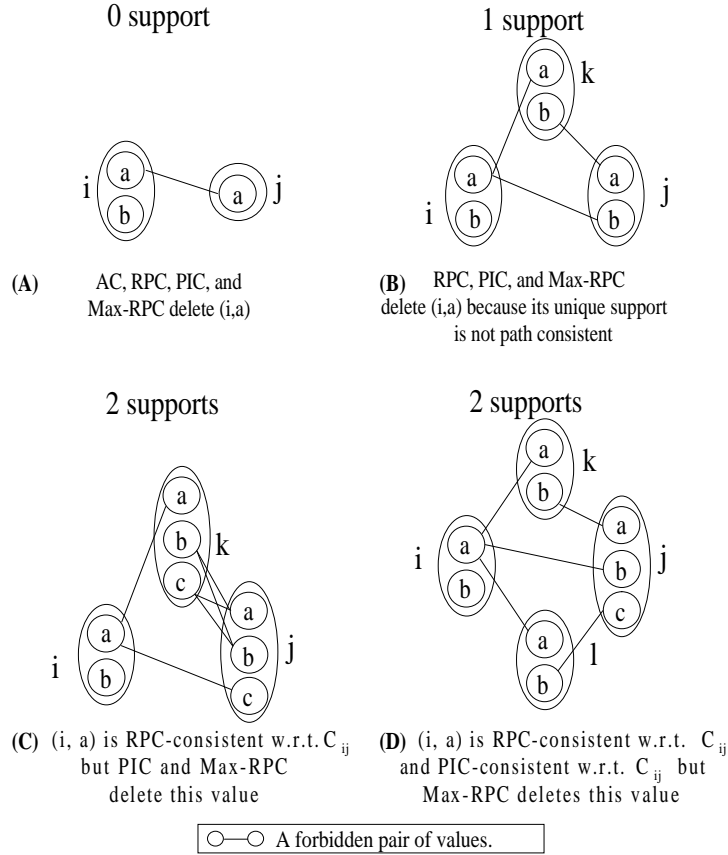

**(A)** AC, RPC, PIC, and Max-RPC delete (i,a)

**(B)** RPC, PIC, and Max-RPC delete (i,a) because its unique support is not path consistent

**(C)** (i, a) is RPC-consistent w.r.t. $C_{ij}$ but PIC and Max-RPC delete this value

**(D)** (i, a) is RPC-consistent w.r.t. $C_{ij}$ and PIC-consistent w.r.t. $C_{ij}$ but Max-RPC deletes this value

○—○ A forbidden pair of values.

Figure 2: Examples showing the relations between PIC, RPC and Max-RPC.

Furthermore, the definition of PIC shows that any constraint network involving less than three variables is path inverse consistent, even though it is not arc consistent.

**Property 1** *A CN is path inverse consistent iff*
- *it involves less than three variables, or*
- *it is arc consistent and for each value (i, a) in $\mathcal{D}$, for any 3-clique $\{i, j, k\}$, (i, a) can be extended to a consistent instantiation of $\{i, j, k\}$.*

This property allows us to see the relations between PIC, RPC and Max-RPC. If a value $(i, a)$ has no support on a constraint $C_{ij}$, the three local consistencies delete this arc inconsistent value (see Figure 2A). If $(i, a)$ has only one support $b$ in $D_j$, PIC, RPC, and Max-RPC delete $(i, a)$ because of $C_{ij}$ if $((i, a), (j, b))$ is path inconsistent (see Figure 2B). The difference between these three local consistencies appears if $(i, a)$ has at least two supports on $C_{ij}$. In such a case, $(i, a)$ is restricted path consistent w.r.t. $C_{ij}$ but PIC can delete it if there is a 3-clique $\{i, j, k\}$ such that all the supports of $(i, a)$ in $D_j$ are path inconsistent because of $k$ (see Figure 2C). So, PIC is more pruningful than RPC. But it often deletes only few additional values because the supports of a value are seldom all path inconsistent because of the same third variable. Max-RPC is far more pruningful since it deletes $(i, a)$ because of $C_{ij}$ if all its supports in $D_j$ are path inconsistent, even if they are not path inconsistent because of the same third variable (see Figure 2D).

## 5. Pruning Efficiency

### 5.1 Qualitative Study

To compare the pruning efficiency of the local consistencies presented above, we use the transitive relation "stronger" introduced in (Debruyne & Bessière, 1997b). A local consistency $LC$ is *stronger* than another local consistency $LC'$ if in any CN in which $LC$ holds, $LC'$ holds too. Consequently, if $LC$ is stronger than $LC'$, any algorithm achieving $LC$ deletes at least all the values removed by an algorithm achieving $LC'$. For instance, since by definition of restricted path consistency RPC is stronger than AC, an RPC algorithm removes at least all the arc inconsistent values. A local consistency $LC$ is *strictly stronger* than another local consistency $LC'$ if $LC$ is stronger than $LC'$ and there is at least one CN in which $LC'$ holds and $LC$ does not.

**Theorem 1** *Restricted path consistency is strictly stronger than AC.*

**Proof** By definition of restricted path consistency, RPC is stronger than arc consistency. Figure 3a shows that there exists a constraint network on which AC holds and RPC does not. Therefore, RPC is strictly stronger than AC.  □

**Theorem 2** *If $k > k' \geq 0$, $k$-RPC is strictly stronger than $k'$-RPC.*

**Proof** The proof that $k$-RPC is stronger than $k'$-RPC if $k > k' \geq 0$ is trivial. Figure 3g shows that there exists a constraint network on which $k'$-RPC holds and $k$-RPC ($k > k' \geq 0$) does not. Therefore, $k$-RPC is strictly stronger than $k'$-RPC if $k > k' \geq 0$.  □

**Theorem 3** *Max-RPC is strictly stronger than $k$-RPC, $\forall k \geq 0$.*

**Proof** The proof that Max-RPC is stronger than $k$-RPC $\forall k \geq 0$ is trivial. Figure 3g shows that for any $k \geq 0$ there exists a constraint network on which $k$-RPC holds and Max-RPC does not. Therefore, Max-RPC is strictly stronger than $k$-RPC $\forall k \geq 0$.  □

**Theorem 4** *If $|\mathcal{X}| \geq 3$, path inverse consistency is strictly stronger than restricted path consistency.*

**Proof** From property 1, PIC is stronger than AC if $|\mathcal{X}| \geq 3$. Now, consider a value $(i, a)$ having only one support $(j, b)$ on $C_{ij}$. If PIC holds, for any third variable $k$, $(i, a)$ can be extended to a consistent instantiation $I$ including $\{i, j, k\}$ and since $b$ is the only support of $(i, a)$ in $D_j$, $I_j = b$. So $((i, a), (j, b))$ is path consistent and $(i, a)$ is restricted path consistent w.r.t. $C_{ij}$. Furthermore, Figure 3b shows that there exists a constraint network on which RPC holds and PIC does not. Therefore, path inverse consistency is strictly stronger than restricted path consistency if $|\mathcal{X}| \geq 3$. □

**Theorem 5** *If $|\mathcal{X}| \geq 3$, max-restricted path consistency is strictly stronger than path inverse consistency.*

**Proof** Suppose there is a max-restricted path consistent CN $P$ with a value $(i, a)$ which is not path inverse consistent. Since the CN is max-restricted path consistent, it is also arc consistent by definition of max-restricted path consistency. Thus, because of property 1 we know there exist two variables $j$ and $k$ such that $\{i, j, k\}$ is a clique in the constraint graph and $(i, a)$ cannot be extended to a consistent instantiation on $\{i, j, k\}$. As a result, none of the supports of $(i, a)$ on $C_{ij}$ is path consistent, which contradicts the assumption that the CN $P$ is max-restricted path consistent. Furthermore, Figure 3i shows that there exists a constraint network on which path inverse consistency hold and max-restricted path consistency does not. Therefore, if $|\mathcal{X}| \geq 3$, max-RPC is strictly stronger □

**Theorem 6** *Singleton arc consistency is strictly stronger than Max-RPC.*

**Proof** Suppose that there exists a CN $P$ with a singleton arc consistent value $(i, a)$ that is not max-restricted path consistent. Let $j \in \mathcal{X}$ be a variable such that $(i, a)$ has no path consistent support in $D_j$. For each support $b$ of $(i, a)$ in $D_j$, there exists a variable $k$ such that $\nexists c \in D_k$ such that $C_{ik}(a, c) \land C_{jk}(b, c)$. Therefore, all the values of $D_j$ are arc inconsistent w.r.t. $P|_{D_i = \{a\}}$ and $(i, a)$ is not singleton arc consistent. So, SAC is stronger than Max-RPC. Figure 3e shows that there exists a constraint network on which Max-RPC holds and SAC does not. Therefore, SAC is strictly stronger than Max-RPC. □

**Theorem 7** *Neighborhood inverse consistency is strictly stronger than max-restricted path consistency.*

**Proof** Let $(i, a)$ be any value of a neighborhood inverse consistent CN $P$. There exists a consistent instantiation $I$ including $i$ and its neighborhood s.t. $I_i = a$. For any $C_{ij} \in \mathcal{C}$, $I_j$ is a path consistent support of $(i, a)$. Indeed, let $k$ be any third variable. If $k$ is linked to $i$, $((i, a), (j, I_j), (k, I_k))$ is a consistent instantiation since $P$ is neighborhood inverse consistent. Otherwise, there are two cases: First, if $k$ is not linked to $j$, $((i, a), (j, I_j), (k, c))$ is consistent $\forall c \in D_k$; Second, if $\exists C_{jk} \in \mathcal{C}$, there exists a consistent instantiation $I'$ of $j$ and its neighborhood s.t. $I'_j = I_j$ and $((i, a), (j, I'_j), (k, I'_k))$ is consistent. So, $(i, a)$ is max-restricted path consistent. Furthermore, Figure 3d shows that there exists a constraint network on which Max-RPC holds and NIC does not. Therefore, NIC is strictly stronger than Max-RPC. □

**Theorem 8** *Strong path consistency is strictly stronger than singleton arc consistency.*

**Proof** Consider a problem that is strong path consistent. Any pair of values can be extended to any third variable. Furthermore, since the problem is strong path consistent, it is also arc consistent and a sub-problem obtained by restricting a domain $D_i$ to a singleton $\{(i, a)\}$ can be made arc consistent. The initial problem is therefore singleton arc consistent. Figure 3c shows that there exists a constraint network on which SAC holds and strong PC does not. Therefore, strong PC is strictly stronger than SAC. $\quad\square$

**Theorem 9** *Singleton restricted path consistency is strictly stronger than singleton arc consistency.*

**Proof** Singleton restricted path consistency is stronger than singleton arc consistency since RPC is stronger than AC. Figure 3d shows that there exists a constraint network on which SAC holds and SRPC does not. Therefore, SRPC is strictly stronger than SAC. $\quad\square$

The stronger relation does not induce a total ordering. Some local consistencies are incomparable.

**Theorem 10**

1. *If $|\mathcal{X}| \geq 3$, path inverse consistency and k-restricted path consistency are incomparable.*

2. *Neighborhood inverse consistency and singleton arc consistency are incomparable.*

3. *Neighborhood inverse consistency and strong path consistency are incomparable.*

4. *Neighborhood inverse consistency and singleton restricted path consistency are incomparable.*

**Proof**

1. cf. Figure 3h and Figure 3j.

2. cf. Figure 3d and Figure 3e.

3. cf. Figure 3d and Figure 3e.

4. cf. Figure 3e and Figure 3f.

Figure 4 summarizes the relations between the local consistencies. There is an arrow from $LC$ to $LC'$ iff $LC$ is strictly stronger than $LC'$. A crossed line between two local consistencies means that they are not comparable w.r.t. the "stronger" relation. When $LC$ is not stronger than $LC'$ ($LC'$ is strictly stronger than $LC$, or $LC$ and $LC'$ are not comparable), a CN in which $LC$ holds and $LC'$ does not can be found in Figure 3. Obviously, the stronger relation is transitive. In Figure 4 we omit the transitivity arcs.

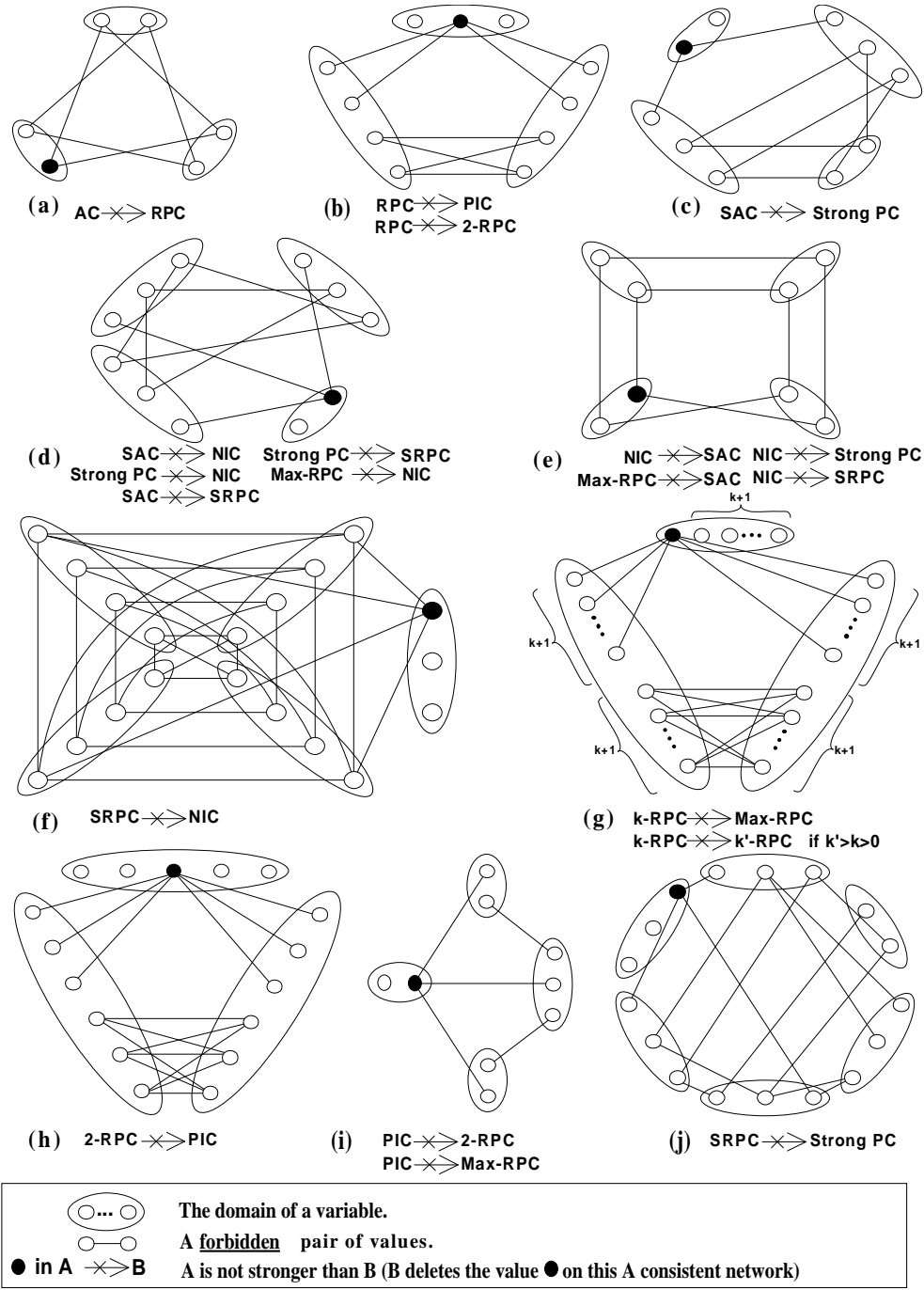

Figure 3: Some CNs proving the "not stronger" relations between some of the mentioned local consistencies.

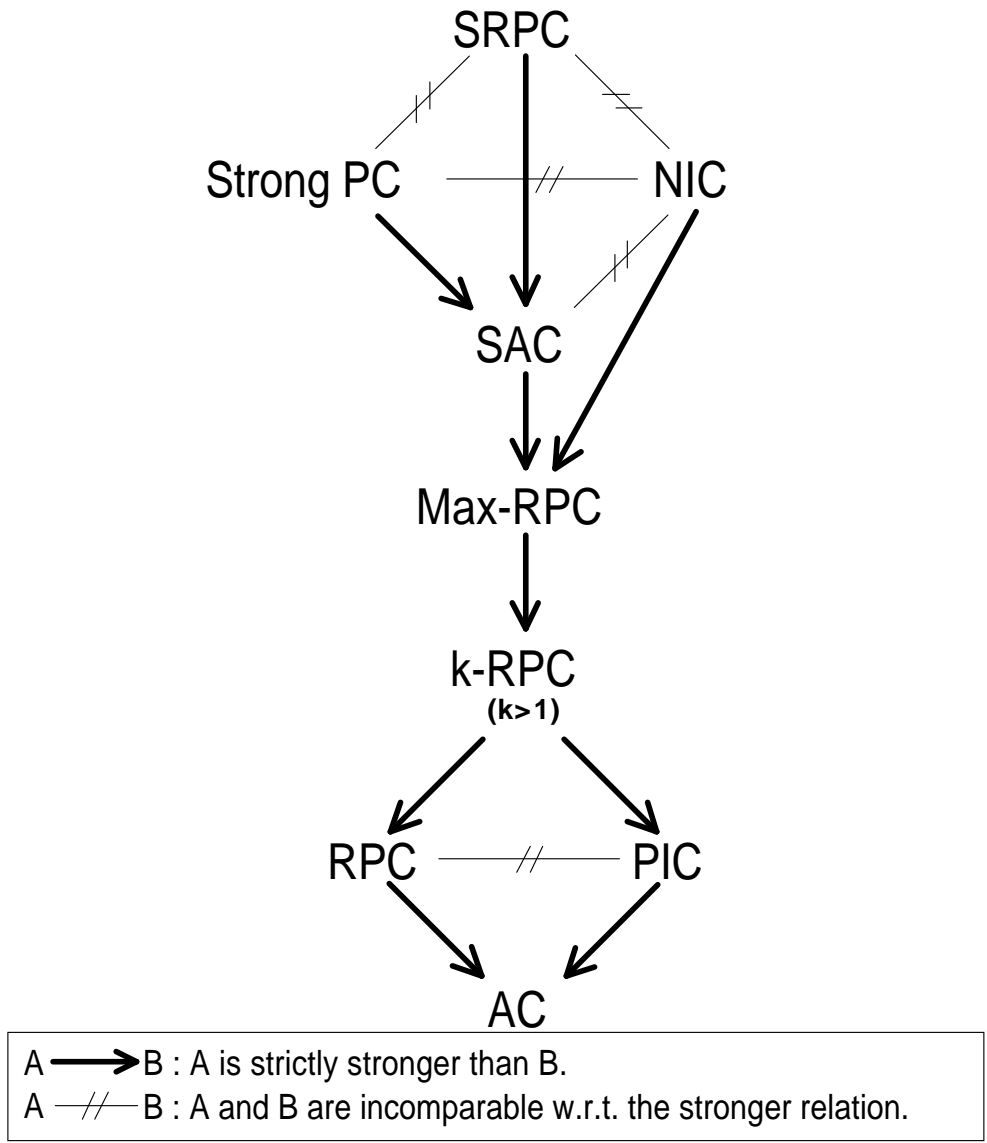

Figure 4: Relations between the mentioned local consistencies.

## 5.2 Experimental Evaluation

Figure 4 does not give any quantitative information. A local consistency $LC$ can remove more values than another local consistency $LC'$ on most of the CNs even though it is incomparable with $LC$ because of some particular CNs. When they are comparable, it does not show if a local consistency is far more pruningful than another or if it performs only few additional value deletions. To have some quantitative information about the pruning efficiency of these local consistencies, we performed an experimental evaluation. The aim of this evaluation is to show how pruningful a local consistency is on random CNs, with a fixed number of variables and values, when the number of constraints and the constraint tightness are changing. We used the random uniform CN generator of (Frost, Bessière, Dechter, & Régin, 1996) which produces instances according to the Model B (Prosser, 1996). It involves four parameters: $n$ the number of variables, $d$ the common size of the initial domains, $p1$ the proportion of constraints in the network (the density $p1=1$ corresponds to the complete graph) and $p2$ the proportion of forbidden pairs of values in a constraint (the tightness). The generated problems have 40 variables and 15 values in each domain. For each local consistency and each density $p1$, two particular values of the tightness have been determined. On the one hand, $T_0(p1)$ is the tightness such that the local consistency does not delete any value on 50% of the CNs generated with $p1$ for density. For values of tightness lower than $T_0(p1)$, the local consistency seldom deletes many values. On the other hand, $T_{all}(p1)$ is the tightness such that the local consistency finds the inconsistency of 50% of the CNs generated with density $p1$. On constraint networks with tighter constraints, the local consistency often removes all the values. For all the mentioned local consistencies, the values $T_0(p1)$ and $T_{all}(p1)$ for any density $p1$ are given in Figure 5 and Figure 6 respectively. We also show these bounds for the variable completability filtering which removes all the globally inconsistent values, and thus is the strongest filtering we can have when we limit filtering to the domains. To determine the $T_0$ and $T_{all}$ bounds, 300 CNs have been generated for each (density, tightness) pair. This explains why the generated problems are relatively small.

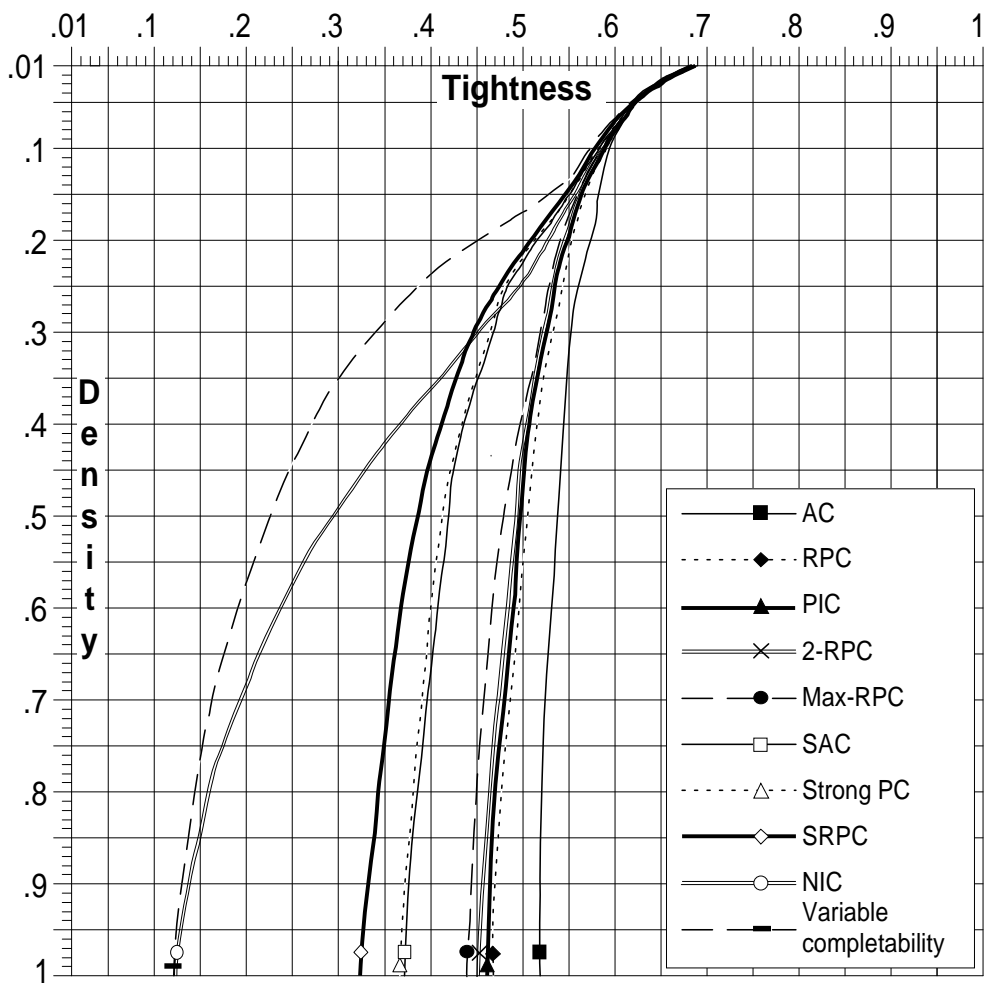

Figure 5: The $T_0$ bounds for random CNs having 40 variables and 15 values in each domain.

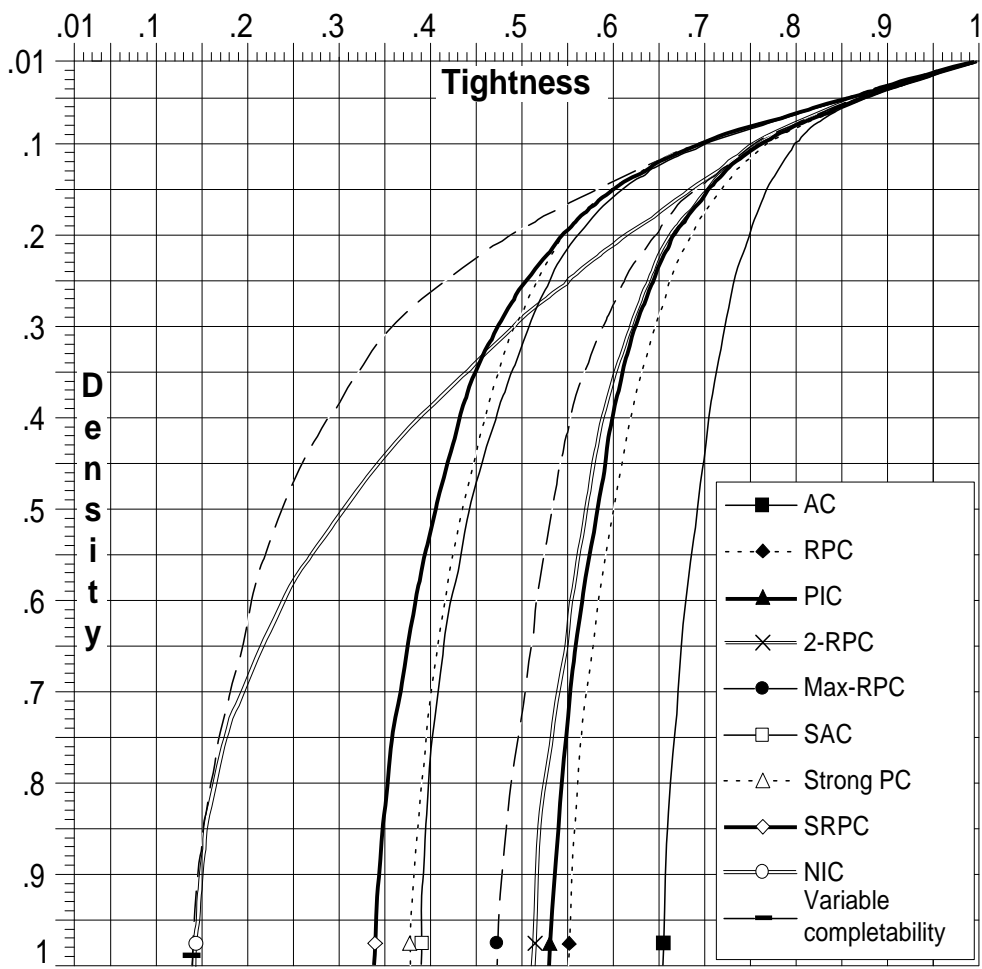

Figure 6: The $T_{all}$ bounds for random CNs having 40 variables and 15 values in each domain.

As already proved theoretically, PIC is stronger than RPC. Their pruning efficiencies are closed. RPC deletes most of the path inverse inconsistent values and is halfway between AC and Max-RPC in terms of pruning efficiency. $k$-RPC with $k > 1$ is incomparable with PIC with regard to the stronger relation. However, Figure 5 and Figure 6 show that 2-RPC is more pruningful than PIC. SAC and strong PC have almost the same pruning efficiency. Their $T_0$ limits merge and their $T_{all}$ limits show a slight difference. This confirms the similitude between SAC and strong PC pointed out in Section 3. Although SRPC and strong PC are not comparable w.r.t. the stronger relation, SRPC removes is more pruningful than strong PC. As predicted in (van Beek, 1994), these polynomial filterings have more difficulties to delete inconsistent values on dense problems with loose constraints. On sparse CNs, the polynomial local consistencies studied are close to variable completability, whereas on very dense CNs, Figure 5 and Figure 6 show a large range of tightnesses between them and variable completability. NIC behaves very differently since on complete constraint networks it corresponds to variable completability. So, on dense CNs, NIC is far more pruningful than the other local consistencies. On CNs generated with a density lower than .28 NIC is less pruningful than SRPC, strong PC and SAC. The more important the propagation through the network is, the closer $T_0$ and $T_{all}$ are. If a filtering (such as AC) uses a very local property to delete inconsistent values, there is a large set of CNs on which it removes some but not all the values. More pruningful local consistencies consider a more important part of the network to know whether a value is consistent or not. So, they seldom delete few values. On most of the CNs, they do not delete any value, or detect inconsistency: the propagation of the first value deletions often leads to a domain wipe out.

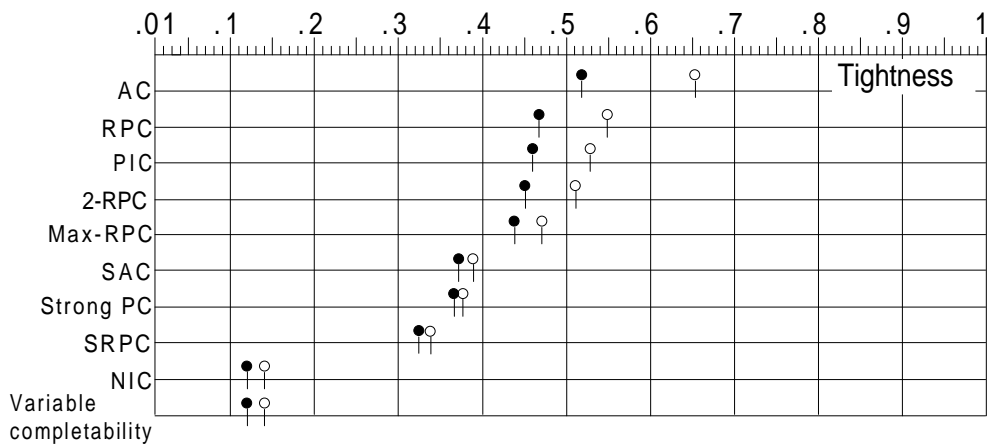

Figure 7: The $T_0$ (black points) and $T_{all}$ (white points) bounds for random CNs having 40 variables, 15 values in each domain, and density 1.

## 6. Time Efficiency

### 6.1 Radio Link Frequency Assignment Problems

An experimental evaluation has been done on the radio link frequency assignment problems described in (Cabon, de Givry, Lobjois, Schiex, & Warners, 1999), namely the instances of the CELAR[4] named Scen01 to Scen11, and the GRAPH instances generated using the GRAPH generator at Delft University named Graph01 to Graph14. In these problems we have to assign frequencies to a set of radio links defined between pairs of sites in order to avoid interferences[5]. These problems have from 200 to 916 variables and there are 40 values in average in each domain. The constraints are binary and have a cost of violation specified by a level from 0 to 4. The level 0 corresponds to hard constraints, and levels from 1 to 4 have a decreasing cost of violation. For each problem ScenXX (resp. GraphXX), we call ScenXX.3, ScenXX.2, ScenXX.1 and ScenXX.0 (resp. GraphXX.3, GraphXX.2, GraphXX.1 and GraphXX.0) the problems of satisfaction obtained by considering the problem ScenXX (resp. GraphXX) with only the constraints of level 0 to 3, 0 to 2, 0 to 1, and 0 respectively.

---

4. We thanks the Centre d'Electronique de l'Armement (France).
5. See http://www-bia.inra.fr/T/schiex/Doc/CELARE.html for a more detailed presentation of these problems.

| | AC7 | RPC2 | PIC2 | Max-RPC1 | SAC1 | SRPC1 | NIC1 |
|--------|------|------|-------|----------|-------|---------|----------|
| Scen02 | 0.27 | 0.7  | 4.38  | 6.33     | 45.5  | 434.93  | 10.45    |
| Scen03 | 0.58 | 1.55 | 9.13  | 14.21    | 99.49 | 946.31  | 26.58    |
| Scen11 | 0.89 | 2.53 | 13.79 | 25.84    | 144.3 | 1362.18 | time out |

Table 2: Cpu time performances on some RLFAP instances on which all the local consistencies studied hold.

In this experimental evaluation, we consider both the cpu time performances and the percentage of values deleted by the local consistencies studied. The algorithms used are AC7 (Bessière, Freuder, & Régin, 1995), RPC2 (Debruyne & Bessière, 1997a), PIC2 (Debruyne, 2000), Max-RPC1 (Debruyne & Bessière, 1997a), the singleton arc consistency algorithm of (Debruyne & Bessière, 1997b) (SAC1) based on AC6, a SRPC algorithm based on RPC2 (SRPC1), and the NIC algorithm proposed in (Freuder & Elfe, 1996) (NIC1) using FC-CBJ (Prosser, 1993) (as in Freuder & Elfe, 1996) with dom+deg dynamic variable ordering heuristic (minimal domain first, in which ties are broken by choosing the variable with the highest degree in the constraint graph Frost & Dechter, 1995; Bessière & Régin, 1996). All these algorithms have been modified to stop as soon as a domain wipe out occurs. We do not show results on strong PC in this section because on these large problems it requires often more than our 2 hours time out limit. These algorithms have been tested on each ScenXX, Scen XX.X, GraphXX, and GraphXX.X problem using a Sun UltraSparc IIi 440 Mhz. For sake of clarity, we only show the results on some representative problems.

### 6.1.1 Results on problems on which all the studied local consistencies hold (cf. Table 2)

If all the local consistencies studied hold on a constraint network, all the corresponding filtering algorithms are useless. They waste time to check whether the local consistencies hold without deleting any inconsistent value. On these problems, the stronger the local consistency is, the more important is the time wasted.

We can see the consequence of the exponential worst case time complexity of NIC1. On most of these problems, NIC1 requires a reasonable cpu time. But as we can see on the problem Scen11, a combinatorial explosion can lead to really prohibitive cpu time for NIC1.

### 6.1.2 Results on arc inconsistent problems (cf. Table 3)

When arc consistency is sufficient to detect the inconsistency of the problem, stronger local consistencies are not always more costly. On Figure 3 we can see that Max-RPC1 has often the best cpu time performances and on Graph06 for example, AC7 is one of the most expensive local consistencies. When enforcing AC requires propagation to find the arc inconsistency of the problem, a stronger local consistency can wipe out a domain more quickly than AC7.

| | AC7 | RPC2 | PIC2 | Max-RPC1 | SAC1 | SRPC1 | NIC1 |
|---|---|---|---|---|---|---|---|
| Scen07 | 0.42 | 0.43 | 0.44 | 0.09 | 0.59 | 0.47 | 1.89 |
| Graph07 | 0.11 | 0.14 | 0.12 | 0.16 | 0.24 | 0.14 | 1.08 |
| Scen08 | 0.75 | 0.48 | 0.73 | 0.4 | 0.52 | 0.47 | time out |
| Graph06 | 0.48 | 0.27 | 0.44 | 0.26 | 0.27 | 0.27 | 10.13 |

Table 3: Cpu time performances on some arc inconsistent RLFAP instances.

| | | AC7 | RPC2 | PIC2 | Max-RPC1 | SAC1 | SRPC1 | NIC1 |
|---|---|---|---|---|---|---|---|---|
| Scen06.1 | cpu time | 0.27 | 0.48 | 0.96 | 2.04 | 66.32 | 227.13 | time out |
| | % of DV | 7.88 | 8.33 | 17.85 | 19.7 | 42.47 | 42.57 | ? |
| Scen09.1 | cpu time | 0.8 | 1.52 | 1.87 | 5.88 | 167.85 | 568.08 | 318.38 |
| | % of DV | 22.48 | 25.79 | 29.79 | 31.03 | 35.86 | 35.86 | 31.57 |
| Graph04 | cpu time | 0.81 | 2.07 | 18.65 | 25.39 | 2238.13 | time out | 101.77 |
| | % of DV | 4.97 | 6.67 | 6.95 | 10.35 | 18.44 | ? | 13.14 |
| Graph10 | cpu time | 1.43 | 3.32 | 37.7 | 51.42 | 3984.13 | time out | 2033.39 |
| | % of DV | 1.43 | 1.62 | 1.68 | 5.42 | 9.53 | ? | 7.35 |
| Graph06.1 | cpu time | 0.39 | 0.81 | 0.9 | 0.8 | 6.69 | 3.21 | 8.54 |
| | % of DV | 14.96 | 17.69 | 100 | 100 | 100 | 100 | 100 |
| Graph12.1 | cpu time | 0.73 | 1.35 | 2.83 | 5.41 | 9.47 | 32.12 | 3.97 |
| | % of DV | 10.42 | 12.23 | 15.28 | 100 | 100 | 100 | 100 |

Table 4: Cpu time performances and percentages of values deleted by the local consistencies studied (% of DV) on some RLFAP instances.

On these constraint networks, all the algorithms used have very low cpu time requirements, except NIC1, which can be very expensive on some instances, such as Scen08.

### 6.1.3 Results on the other problems (cf. Table 4)

On many of the RLFAP problems the local consistencies do not delete the same sets of inconsistent values. We can see an important difference between the pruning efficiencies especially on the problems ScenXX.1 and GraphXX.1.

Obviously, on most of these problems, the more pruningful the local consistency is, the more important is the time required. We can see this on the problems Scen06.1 and Scen09.1 for example. However, AC7, RPC2, PIC2, and Max-RPC1 have cpu time performances in the same order of magnitude while SAC1, SRPC1, and NIC1 are often far more expensive.

This is especially obvious on Graph04 and Graph10. However, it is difficult to say which is the most interesting local consistency on these problems since even if SAC1, and SRPC1 are costly, we can see on Scen06.1 and Graph04 that they can be far more pruningful.

These problems highlight that NIC1 is not very stable. It sometimes shows good performances, but an exponential explosion can lead to a prohibitive cost on some instances. When NIC1 requires a reasonable time, its pruning efficiency is closer to the one of Max-RPC1 than to the one of SAC1. These results confirm that if the neighborhoods of the variables are not small, NIC1 can be really prohibitive.

On Graph06.1, PIC2 (and obviously the algorithms enforcing a stronger local consistency) finds the inconsistency of the problem whereas AC7, and RPC2 remove only a part of the inconsistent values. We can see a similar behavior on Graph12.1 where Max-RPC1 wipes out a domain whereas AC7, RPC2 and PIC2 do not find the inconsistency of the problem. On these instances, Max-RPC1 is the best choice.

## 6.2 Randomly Generated Problems

The random uniform CN generator of section 5.2 is used to compare the cpu time required to enforce the local consistencies. We have to point out that NIC has not been designed to be used on uniform CNs but to adapt filtering effort to the degree of the variables in the constraint graph. So, NIC would have better performances on non-uniform CNs than those presented in this section. The generated problems have 200 variables and 30 values in each initial domain. Figure 8 shows the results on CNs with density of .02. These CNs are relatively sparse since the variables have four neighbors on average. Figure 9 presents performances at density .15 (the variables have 30 neighbors on average). Because of the set of parameters, there are no flawed variables (MacIntyre, Prosser, Smith, & Walsh, 1998) in the generated problems.[6] In addition to the algorithms of the previous section, we use a strong path consistency algorithm based on PC8 (Chmeiss & Jégou, 1996) and AC6. This algorithm stops as soon as a domain wipe out occurs or as soon as a constraint no longer allows any pair of values. In addition to the percentage of deleted values and cpu time performances, Figure 8 and Figure 9 show the cpu time to number of deleted values ratio for each tightness where the local consistency removes at least one value on average. For each tightness, 50 instances were generated. Figure 8 and Figure 9 show mean values obtained on a Pentium II-266 Mhz with 32 Mb of memory under Linux.

As observed in (Gent, MacIntyre, Prosser, Shaw, & Walsh, 1997) for arc consistency, the filtering algorithms tested have a complexity peak. For low values of the tightness, they easily prove that the values are locally consistent, and when constraints are very tight, they quickly wipe out a domain. Each local consistency has a phase transition where most of the hardest problems for an algorithm achieving this local consistency tend to occur.

## 6.3 Experiments on Sparse CNs

Even on sparse CNs (see Figure 8), the cpu time results are so different between the algorithms (7h 48min for strong PC at its peak when AC7 requires at most .22 seconds on average) that a logarithmic scale has to be used. Strong PC is really prohibitive, even for low values of tightness. SRPC and SAC have bad cpu time to number of deleted values ratios, except SAC on CNs having very tight constraints because the SAC algorithm used is based on AC6 which can be more efficient than AC7 on such problems. On these sparse CNs, NIC has often better cpu time performances than SAC but it does not remove more values than Max-RPC. Consequently, NIC has a bad cpu time to number of deleted values ratio. Unlike strong PC, SRPC, SAC, and NIC, the cpu time requirements of AC7, PIC2, RPC2 and Max-RPC are of the same order of magnitude. The cpu time to number of deleted values ratios of these four last filterings are also very close, with a little advantage for PIC2. Although PIC is stronger than RPC, PIC2 can be less expensive than RPC2 on sparse CNs. If there are few 3-cliques in the constraint graph, PIC2 does not require far more cpu time than AC7 whereas RPC2 is about two times as expensive as AC7 since it looks for two supports for each value on each constraint.

---

6. In Section 5.2, the tightness reaching 1, there was obviously flawed variables for some sets of parameters.

## 6.4 Experiments on more Dense CNs

On more dense CNs (see Figure 9), the complexity peaks of AC7, RPC2, PIC2, and Max-RPC stay close to each other. PIC2 is less worthwhile since it deletes few additional values compared to RPC2 while its cpu time requirements are close to those of Max-RPC. Max-RPC has one of the best cpu time to number of deleted values ratios. As soon as RPC leads to a domain wipe out, the cpu time performances of SRPC and RPC2 merge. Indeed, the SRPC algorithm used enforces RPC2 before checking the restricted path consistency of the sub-problems $P|_{D_i = \{a\}}$ for each $(i, a) \in \mathcal{D}$. If all the values of a domain are restricted path inconsistent, the RPC preprocessing finds the global inconsistency of the problem and the SRPC algorithm stops. SRPC is less expensive than strong PC although it is more pruningful. These two filterings remain the most expensive. NIC is the most pruningful local consistency on these CNs. Hence, on a large range of tightnesses, NIC has the best cpu time to number of deleted values ratio. However, on some instances, NIC cannot avoid the combinatorial explosion. Although NIC requires "only" fifteen minutes on average at tightness .52, more than two hours are required on some instances. It is conceivable that instances on which NIC requires far more cpu time exist for this set of parameters. Obviously, the set of CNs on which NIC is prohibitive grows when the density increases. The results on SAC have a lower standard deviation. SAC never requires more than fifty two minutes on the problems generated for these experiments.

## 6.5 Discussion

What can we conclude from these results? Strong PC is by far the least interesting filtering technique. Compared to SAC, which removes most of the strong path inconsistent values, strong PC is really prohibitive.[7] Achieving SAC or SRPC is costly as long as these two local consistencies do not delete any value. Obviously, although SAC and SRPC are more expensive than Max-RPC on almost all the generated problems, we cannot say that it is better to use Max-RPC. Indeed, at density .15 for example, Max-RPC is useless for tightnesses lower than .63 since it does not delete any value, while for SRPC the limit is .57 of tightness. Furthermore, for singleton consistencies we can argue that the algorithm used to achieve them is not optimal. An algorithm reusing part of the filtering performed on $P|_{D_i=\{a\}}$ to process other sub-problems $P|_{D_j=\{b\}}$, $((i,a)$ and $(j,b)$ belonging to $\mathcal{D})$ would improve cpu time performances. However, the cpu time to number of deleted values ratios of SAC and SRPC algorithms are often among the worst ones, especially on sparse CNs. SAC and SRPC are so expensive that it is hardly likely that enhancements of these algorithms could lead them to be the most worthwhile filterings. On sparse uniform CNs, NIC is not the best choice. Compared to Max-RPC, it does not delete enough values to offset the additional cpu time cost. Furthermore, NIC cannot be used on dense CNs since its cpu time requirements become greater than those of a search algorithm. So, NIC has to be used only on "relatively" dense CNs, as those of Figure 9 on which NIC is worthwhile on average (although on some instances a combinatorial explosion cannot be avoided). On very dense CNs, the worst case time complexity of Max-RPC and PIC2 is close to the one of the best path consistency algorithm ($O(en + ed^2 + cd^3)$ against $O(n^3d^3)$). However, the experiments underline that achieving Max-RPC and PIC2 is far less expensive in practice. Compared to RPC2 and Max-RPC, PIC2 is not a good solution in-between. The cpu time to number of deleted values ratios of RPC2 and Max-RPC are better than the one of PIC2 (except on very sparse CNs on which PIC2 can be less expensive than RPC2). Indeed, PIC2 deletes only few additional values compared to RPC2, while its cpu time performances are close to those of Max-RPC.

---

7. We can point out that when the path consistency of a constraint can be expressed without explicitly storing the set of forbidden tuples, path consistency can be used (e.g., temporal networks Allen, 1983, constraint networks Smith, 1992).

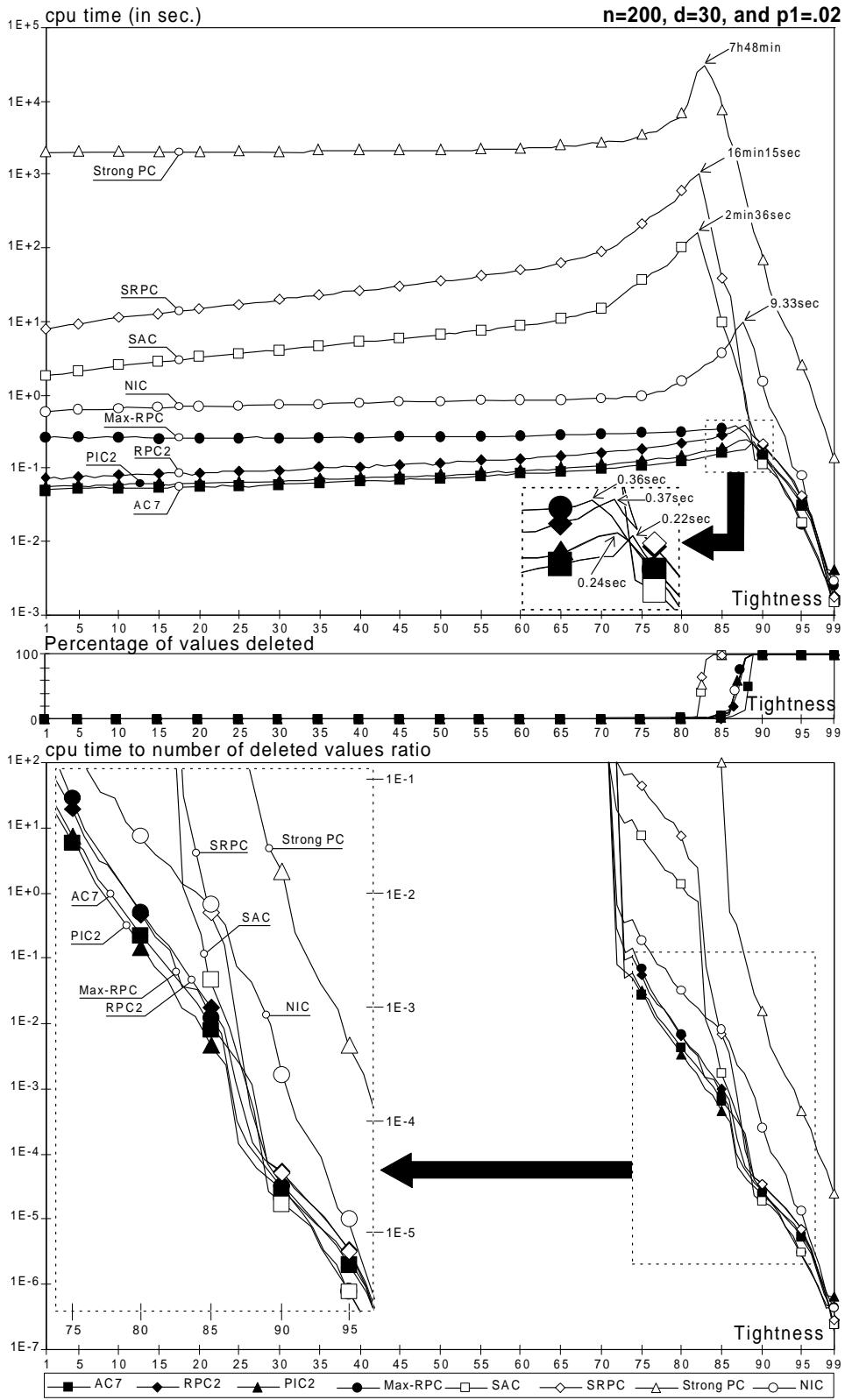

Figure 8: Experimental evaluation on random CNs with $n=200$, $d=30$, and $p1=.02$.

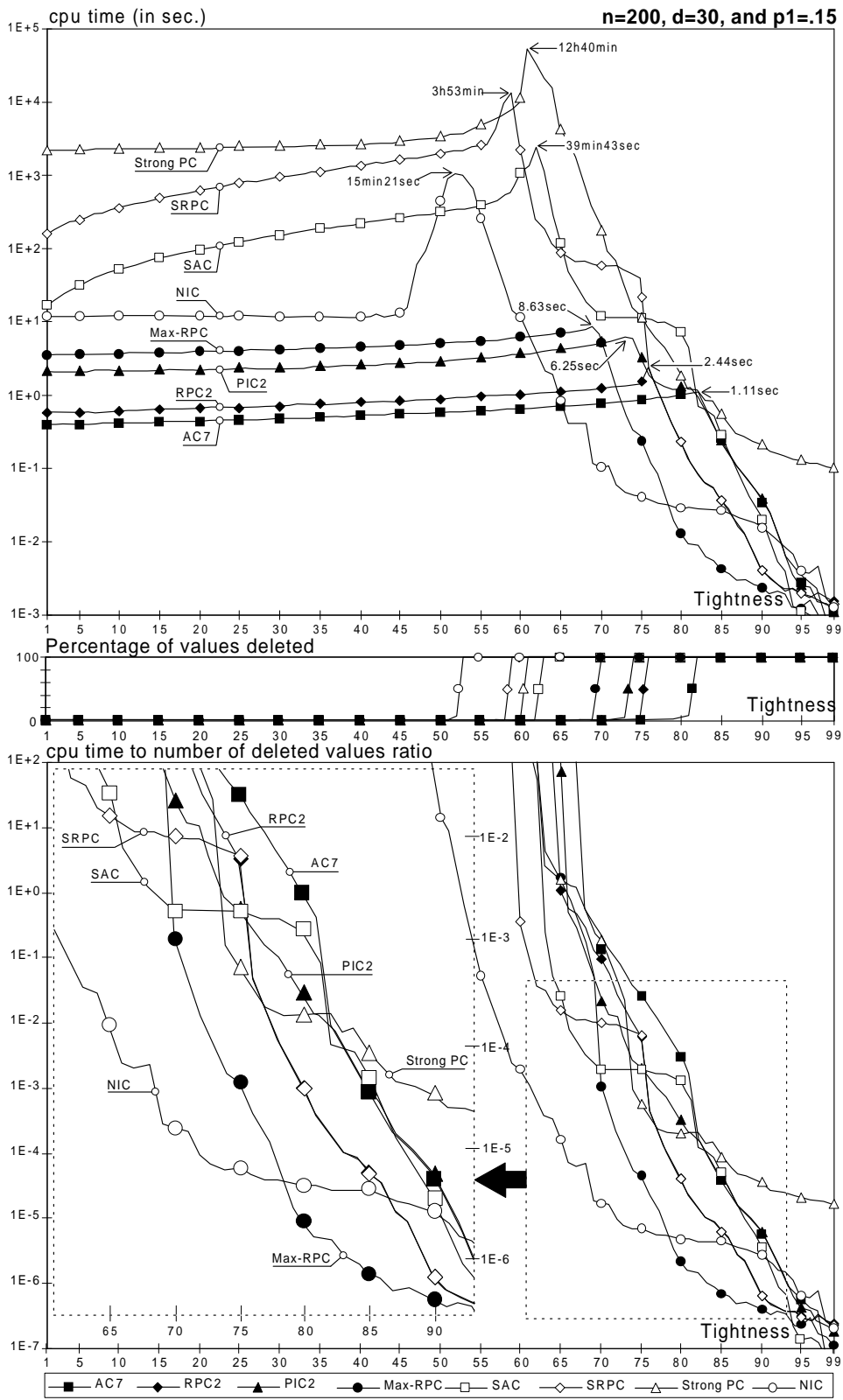

Figure 9: Experimental evaluation on random CNs with $n=200$, $d=30$, and $p1=.15$.

Cpu time performances are even more essential when the aim is to maintain a local consistency during search. Maintaining a local consistency during search requires to repeatedly propagate the choice of a value for a variable (namely the restriction of a domain to a singleton) or the refutation of a value. To be worthwhile, a local consistency has to require less time to detect that a branch of the search tree does not lead to a solution than a search algorithm to explore this branch. So, maintaining a local consistency during search can outperform MAC on hard problems only if this local consistency is more pruningful than AC while requiring only a little additional cpu time. With regard to this criterion, we can discard strong PC, SAC, SRPC, and NIC on dense CNs because they are too expensive. It is conceivable that we can find instances on which maintaining these consistencies during search outperforms MAC, but the more expensive the maintained local consistency is, the more seldom the problems on which MAC is outperformed will be. On sparse CNs, NIC is not prohibitive, but it deletes only few additional values compared to Max-RPC and it has therefore a bad cpu time to number of deleted values ratio. Finally, The most promising local consistencies are RPC and Max-RPC. If we exclude arc consistency, RPC is the least expensive local consistency we studied. Furthermore, the RPC algorithms delete most of the path inverse inconsistent values. Although Max-RPC is far more pruningful than arc consistency, experiments show that in practice, Max-RPC has very good cpu time results. Therefore, it seems very likely that maintaining RPC or Max-RPC during search could outperform MAC on very hard problems.

To confirm these results, an algorithm called Quick that maintains an adaptation of Max-RPC has been compared to MAC. The results of these experiments (Debruyne, 1999) show that Quick has better cpu time performances than MAC on large and hard randomly generated CNs that are relatively sparse. More interestingly, Quick has a more impor-tant stability than MAC (the cpu time performances of Quick have a very low standard deviation). It would be very interesting to propose efficient algorithms that maintain the local consistencies studied in this paper and to compare these algorithms. Such a study would allow us to know whether during search, the more advantageous local consistencies remain RPC and Max-RPC as during a preprocessing step. First results on the effect of maintaining SAC during search are given in (Prosser, Stergiou, & Walsh, 2000).

## 7. Conclusion

In this paper we extended the idea of restricted path consistency to $k$-RPC and Max-RPC, which are more pruningful local consistencies. We proposed a new class of local consistencies called singleton consistencies. We studied these new local consistencies and the other local consistencies that alike can be used on large CNs while removing more values than arc consistency. We showed some relations between them and we compared both theoretically and experimentally their pruning and time efficiencies. The most pruningful are neighborhood inverse consistency and singleton restricted path consistency. However, SRPC is expensive in time and the exponential worst case time complexity of NIC makes it unusable on dense CNs. If we are looking for a local consistency that would advantageously be maintained during search, RPC and Max RPC seem to be the most promising local consistencies. Indeed, they are relatively inexpensive and far more pruningful than arc consistency.

## 8. Acknowledgements

We would like to thank Toby Walsh for his suggestions for improving the presentation of the figures in Section 5.

## References

Allen, J. (1983). Maintaining Knowledge about Temporal Intervals. *Communications of the ACM, 26(11)*, 832–843.

Bacchus, F., & van Run, P. (1995). Dynamic variable ordering in csps. In *Proceedings of CP-95, Cassis, France*, pp. 258–275.

Berlandier, P. (1995). Improving domain filtering using restricted path consistency. In *Proceedings of IEEE CAIA-95*.

Bessière, C., Freuder, E., & Régin, J. (1995). Using inference to reduce arc-consistency computation. In *Proceedings of IJCAI-95, Montréal, Canada*, pp. 592–598.

Bessière, C., Freuder, E., & Régin, J. (1999). Using constraint metaknowledge to reduce arc consistency computation. *Artificial Intelligence, 107(1)*, 125–148.

Bessière, C., & Régin, J. (1996). MAC and combined heuristics: Two reasons to forsake FC (and CBJ?) on hard problems. In *Proceedings of CP-96, Cambridge MA*, pp. 61–75.

Cabon, C., de Givry, S., Lobjois, L., Schiex, T., & Warners, J. (1999). Radio link frequency assignment benchmarks. *CONSTRAINTS, 4(1)*, 79–89.

Cheeseman, P., Kanefsky, B., & Taylor, W. (1991). Where the really hard problems are. In *Proceedings of IJCAI-91, Sydney, Australia*, pp. 294–299.

Chmeiss, A., & Jégou, P. (1996). Two new constraint propagation algorithms requiring small space complexity. In *Proceedings of IEEE ICTAI-96, Toulouse, France*, pp. 286–289.

Debruyne, R. (1999). A strong local consistency for constraint satisfaction. In *Proceedings of IEEE ICTAI-99, Chicago IL*, pp. 202–209.

Debruyne, R. (2000). A property of path inverse consistency leading to an optimal algorithm. In *Proceedings of ECAI-00, Berlin, Germany*, pp. 88–92.

Debruyne, R., & Bessière, C. (1997a). From restricted path consistency to max-restricted path consistency. In *Proceedings of CP-97, Linz, Austria*, pp. 312–326.

Debruyne, R., & Bessière, C. (1997b). Some practicable filtering techniques for the constraint satisfaction problem. In *Proceedings of IJCAI-97, Nagoya, Japan*, pp. 412–417.

Dechter, R., & Meiri, I. (1994). Experimental evaluation of preprocessing algorithms for constraint satisfaction problems. *Artificial Intelligence, 68*, 211–241.

Dechter, R., & Pearl, J. (1988). Network-based heuristics for constraint-satisfaction problems. *Artificial Intelligence, 34*, 1–38.

Freuder, E. (1982). A sufficient condition for backtrack-free search. *Journal of the ACM, 29(1)*, 24–32.

Freuder, E. (1985). A sufficient condition for backtrack-bounded search. *Journal of the ACM, 32(4)*, 755–761.

Freuder, E. (1991). Completable representations of constraint satisfaction problems. In *Proceedings of KR-91, Cambridge MA*, pp. 186–195.

Freuder, E., & Elfe, C. (1996). Neighborhood inverse consistency preprocessing. In *Proceedings of AAAI-96, Portland OR*, pp. 202–208.

Frost, D., Bessière, C., Dechter, R., & Régin, J. (1996). Random uniform csp generators. In *http://www.ics.uci.edu/˜ frost/csp/generatotr.html*.

Frost, D., & Dechter, R. (1995). Look-ahead value ordering for constraint satisfaction problems. In *Proceedings of IJCAI-95, Montréal, Canada*, pp. 572–578.

Gaschnig, J. (1974). A constraint satisfaction method for inference making. In *Proceedings of the 12th Annual Allerton Conf. Circuit System Theory, U.I.L., Urbana-Champaign IL*, pp. 866–874.

Gent, I., MacIntyre, E., Prosser, P., Shaw, P., & Walsh, T. (1997). The constrainedness of arc consistency. In *Proceedings of CP-97, Linz, Austria*, pp. 327–340.

Golomb, S., & Baumert, I. (1965). Backtrack programming. *Journal of the ACM, 12(4)*, 516–524.

Grant, S., & Smith, B. (1996). The phase transition behaviour of maintaining arc consistency. In *Proceedings of ECAI-96, Budapest, Hungary*, pp. 175–179.

Haralick, R., & Elliott, G. (1980). Increasing tree search efficiency for constraint satisfaction problems. *Artificial Intelligence, 14*, 263–313.

Kumar, V. (1992). Algorithms for constraint satisfaction problems: A survey. *AI Magazine, 13(1)*, 32–44.

MacIntyre, E., Prosser, P., Smith, B., & Walsh, T. (1998). Random constraint satisfaction: theory meets practice. In *Proceedings of CP-98, Pisa, Italy*, Vol. 19, pp. 325–339.

McGregor, J. (1979). Relational consistency algorithms and their application in finding subgraph and graph isomorphisms. *Information Sciences, 19*, 229–250.

Meseguer, P. (1989). Constraint satisfaction problems: An overview. *AICOM, 2*, 3–17.

Nadel, B. (1988). Tree search and arc consistency in constraint satisfaction algorithms. *in L. Kanal and V. Kumar, editors, Search in Artificial Intelligence, Springer-Verlag*, 287–342.

Prosser, P. (1993). Hybrid algorithms for the constraint satisfaction problem. *Computational Intelligence, 9(3)*, 268–299.

Prosser, P. (1996). An empirical study of phase transition in binary constraint satisfaction problems. *Artificial Intelligence, 81*, 81–109.

Prosser, P., Stergiou, K., & Walsh, T. (2000). Singleton consistencies. In *Proceedings of CP-00, Singapore*, pp. 353–368.

Sabin, D., & Freuder, E. (1994). Contradicting conventional wisdom in constraint satisfaction. In *Proceedings of ECAI-94, Amsterdam, Netherlands*.

Schiex, T., Régin, J., Gaspin, C., & Verfaillie, G. (1996). Lazy arc consistency. In *Proceedings of AAAI-96, Portland OR*, pp. 216–221.

Singh, M. (1995). Path consistency revisited. In *Proceedings of IEEE ICTAI-95, Washington D.C.*

Smith, B. (1992). How to Solve the Zebra Problem, or Path Consistency the Easy Way. In *Proceedings of ECAI-92*, pp. 36–37.

Tsang, E. (1993). *Foundations of Constraint Satisfaction*. London, Academic Press.

van Beek, P. (1994). On the inherent level of local consistency in constraint networks. In *Proceedings of AAAI-94, Seattle WA*, pp. 368–373.